\documentclass[letterpaper, 10 pt, conference]{ieeeconf}  

\IEEEoverridecommandlockouts                              
\usepackage{amsmath} 
\usepackage{amssymb}  
\usepackage{algorithm}
\usepackage[noend]{algpseudocode}
\usepackage{graphicx}
\usepackage{subcaption}
\usepackage{multirow}
\usepackage{array}
\usepackage{url}
\usepackage[capitalize]{cleveref}
\usepackage{booktabs,siunitx}
\usepackage[keeplastbox]{flushend}
\title{\LARGE \bf
Layer-wise synapse optimization for implementing neural networks on general neuromorphic architectures 
}

\author{John Mern,$^{1}$ Jayesh K. Gupta,$^{2}$ and Mykel J. Kochenderfer$^{3}$
\thanks{*This work was supported by the Army Research Laboratory}
\thanks{$^{1}$J. Mern is Graduate Student in the Stanford Intelligent Systems Laboratory, 
		Department of Aeronautics and Astronautics, Stanford University}%
\thanks{$^{2}$J. K. Gupta is Graduate Student in the Stanford Intelligent Systems Laboratory, 
		Department of Computer Science, Stanford University}%
\thanks{$^{3}$M. J. Kochenderfer is Assistant Professor in Department of Aeronautics and Astronautics, Stanford University}%
}

\begin{document}

\maketitle
\thispagestyle{empty}
\pagestyle{empty}

\begin{abstract}

Deep artificial neural networks (ANNs) can represent a wide range of complex functions. Implementing ANNs in Von Neumann computing systems, though, incurs a high energy cost due to the bottleneck created between CPU and memory. Implementation on neuromorphic systems may help to reduce energy demand. Conventional ANNs must be converted into equivalent Spiking Neural Networks (SNNs) in order to be deployed on neuromorphic chips. 
  This paper presents a way to perform this translation. We map the ANN weights to SNN synapses layer-by-layer by forming a least-square-error approximation problem at each layer. 
  An optimal set of synapse weights may then be found for a given choice of ANN activation function and SNN neuron. Using an appropriate constrained solver, we can generate SNNs compatible with  digital, analog, or hybrid chip architectures. We present an optimal node pruning method to allow SNN layer sizes to be set by the designer.
To illustrate this process, we convert three ANNs, including one convolutional network, to SNNs. 
 In all three cases, a simple linear program solver was used. The experiments show that the resulting networks maintain agreement with the original ANN and excellent performance on the evaluation tasks. 
 The networks were also reduced in size with little loss in task performance. 

\end{abstract}

\section{INTRODUCTION}

	The ability of deep artificial neural networks (ANNs) to represent a broad class of complex functions has made them especially useful for machine learning applications. Currently, most implementations of ANNs in machine learning communicate in real-valued signals of 32-bit or higher resolution and are therefore incompatible with neuromorphic systems, which communicate in discrete 1-bit pulses. Further challenges are introduced by constraints imposed by specific neuromorphic chip sets, such as limits on weight range and resolution. Spiking Neural Networks (SNNs) are a class of ANN that uses spiking activation functions at the neural layers and are therefore compatible with neuromorphic implementations~\cite{Maass1997}. Gradient-based training methods, commonly used on conventional ANNs, cannot be directly applied to the inherently discontinuous spiking neurons in SNNs without incurring loss. Instead, a common approach is to first train an ANN via stochastic-gradient-descent (SGD) and then convert the learned network weights to weights appropriate for an SNN~\cite{Hunsberger2016}. Existing translation methods are limited in the architectures to which they may be applied. 
    
	Neuromorphic chips have the potential to significantly reduce the power requirements of deep neural network implementations. In conventional Von Neumann hardware, the highly parallel operations of neural network calculation tend to create a bottleneck between the CPU and memory~\cite{Indiveri2015}. As a result, most deep networks are extremely energy inefficient when run on these architectures~\cite{Li2016}. Neuromorphics alleviate the bottleneck by co-locating memory and processing in a single neuro-synaptic mesh; synaptic channel weights act as memory and neurons handle computation.  Additionally, because the neurons operate through spikes, energy is only expended when needed for computation, resulting in overall lower power consumption for parallel operations~\cite{Boahen2017}. Neuromorphics may then offer a path to deploy ANNs on power-constrained devices that would otherwise be infeasible.

	ANNs have been successfully deployed to neuromorphic chips in a few limited cases. Most existing methods depend on directly mapping learned ANN weights to SNNs, subject to constraints~\cite{Diehl2016,Esser2016}. A significant drawback for these types of methods is that they require the SNN neuron firing rate curves to closely emulate the ANN activation function. Most analog and hybrid neuromorphic chips use a form of leaky integrate and fire (LIF) neuron~\cite{Orhan2012}, which are highly non-linear and have a gradient approaching infinity near the voltage threshold. Methods relying on weight mapping are therefore only applicable to digital neuromorphic architectures in which the simulated neuron function can be artificially controlled~\cite{Diehl2016a}.
    Newly emerging neuromorphic architectures are mixed analog-digital, which use analog neurons. These hybrid chips offer theoretically lower power consumption, making a more general translation approach desirable~\cite{Benjamin2014}. Additionally, existing approaches often replicate the network multiple times to increase accuracy, requiring a large number of neurons.

	The method introduced here first trains an ANN using a gradient-based approach and then translates it into an SNN with a similar architecture. The method proposed, which we call layer-wise synapse optimization (LSO), translates the ANN layer-by-layer, solving for the synaptic weights such that the hidden-layer activations of the ANN are optimally represented in the SNN. This method formulates the translation problem as a linear least-square error problem at each layer, accounting for the SNN neuron behavior. This approach allows the SNN to optimally represent the ANN features for a given pair of ANN activation function and SNN neuron, though analogous pairs (i.e. rectified linear unit (ReLU) and LIF) would be expected to retain a better mapping. Introduction of noise at each translation helps to account for the spiking behavior which is not captured in the firing rate-approximation. 
    
    Because the method operates on pre-trained ANNs, existing state-of-the-art networks (i.e. AlexNet, VGG, etc.) 
    can be employed without replicating the data-intensive training process. Additionally, networks for tasks requiring specialized, often difficult to implement training approaches, may be trained using proven methods without additional constraint. This would be especially beneficial for fields such as reinforcement learning.
    
    The method also introduces an optimal compression method, which allows the SNN layer size to be selected independent of the corresponding ANN layer size. In this way, architecture-specific constraints on neuron-ensemble or core size may be imposed with minimal loss. In contrast with previous works, LSO does not require the SNN to have identical or linearly replicated structure to the source ANN. In addition, it does not require the ANN non-linearities to emulate the neuron spike-rate behavior, making it applicable to hybrid and analog architectures in addition to digital. It can be applied to feed-forward multi-layer perceptron (MLP) networks including convolutional neural networks (CNN). 

\section{METHOD} \label{sec: Method}
	This work proposes a translation approach that translates the hidden-layer activations of the ANN to SNN representation layer-by-layer. Instead of mapping ANN weights directly onto the SNN synapses, the optimal weights are identified by fitting SNN neuron outputs to ANN activation-function outputs for a given sample input set. 
    
\begin{figure}
\includegraphics[width=0.5\textwidth]{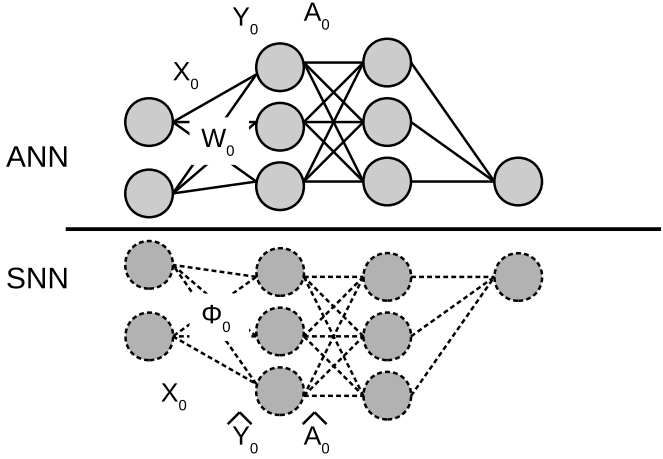}
\caption{Example neural network with feature labels}
\label{fig:NN}
\end{figure}
    
	At each layer $i$, the algorithm solves for the post-neuron synaptic weights, or decoders, $\phi_i$ such that the features of the ANN layer $A_i$ are optimally represented by the equivalent SNN layer. \Cref{fig:NN} shows an MLP with the network values labeled with the notation used in this section. $X_i$ are the prior layer outputs (or network inputs for the first layer) and $Y_i$ are the ANN layer inputs generated through the pre-multiplication of $X_i$ with $W_i$. In the SNN, $\hat{Y_i}$ are the neural layer inputs of the SNN that approximate $Y_i$ for the ANN, and $\hat{A_i}$, are the analogous activations. The hidden layer solution process is outlined in~\crefrange{eq:start}{eq:end}:
   \begin{align}
    Y_i &= W_iX_i \label{eq:start}\\
    A_i &= f(Y_i) \label{eq:ann_act}\\
    A_i &= A_i * scale \\
    \hat{Y_i} &= \Call{InvLIF}{A_i} \label{eq:inlif}\\
    \tilde{Y_i} &= \Call{Truncate}{\hat{Y_i},s} \\
    \phi_i &= (X_i)^+\tilde{Y_i} \\
    \hat{A_i} &= \phi_iX_i \\
    X_{i+1} &= \hat{A_i}+\epsilon\label{eq:end}
    \end{align}
    The process is described in the remainder of this section. 
   
   LSO works by computing what the input to the SNN layer must be such that the output of that layer will match the output of the ANN activations. 
    A set of samples is required to generate the target feature set for each layer. 
    This sample set should be chosen such that it provides good coverage of the input space of interest for the network.
    First, the algorithm replicates the provided samples a selected number of times. These repeated samples will be used to increase sensitivity to neural noise, as described later in this section. 
    The repeated samples are used for the first layer as $X_0$. 
    
    The input samples are passed to the first ANN layer activation-function, which generates the layer features $A_i$, as shown in equations~\cref{eq:start} and~\cref{eq:ann_act}.
    These features are scaled such that the maximum feature activation does not exceed the network's maximum firing rate.
    This step was inspired by previous work~\cite{Diehl2015}.
    Once the ANN features are known, they cannot be directly compared to the SNN layer outputs as the SNN neurons operate in a time-domain, whereas the ANN does not.
    In order to use this method, we approximate a constant-time SNN output as the neuron firing rate to give an approximate estimate of the average neuron output.
    For the basic Leaky-Integrate-and-Fire (LIF) neuron used in this study, a firing rate equation and its inverse may be easily derived. This is shown below for a neuron with input current $I(t)$, membrane time-constant $\tau_m$, membrane resistance $R$, and voltage threshold $v_\text{th}$. 
    
    We can begin by considering the differential equation for membrane voltage $v$:
\begin{equation} \tau_m \frac{dv}{dt} = -v(t) + RI(t)\end{equation}
Assuming constant input $I$ over integration time-step 
\begin{equation} v(t) = RI(1-\exp(-\frac{t}{\tau_m}))\end{equation}
The neuron spikes when $v(t)$ exceeds a threshold voltage $v_\text{th}$. The spike period $T$ is 
\begin{equation} T = \tau_m \ln\left(\frac{RI}{RI-v_\text{th}}\right) \end{equation}
Adding a refractory period $\Delta_\text{ref}$ and inverting gives the firing frequency $f$
\begin{equation}f=\left(\Delta_\text{ref} + \tau_m \ln\left(\frac{RI}{RI-v_\text{th}}\right)\right)^{-1} \end{equation}
Finally, the inverse of this function with respect to the input current is 
\begin{equation} I = \Call{InvLIF}{f} = \frac{e^Cv_\text{th}}{R(e^C-1)}\end{equation}
\begin{equation} C = \frac{f^{-1}-\Delta_\text{ref}}{\tau_m} \end{equation}

The domain and range of this \Call{InvLIF} equation is limited to $\{x \mid x\geq0\}$.
Strict implementation of this would result in all negative ANN inputs at $Y_i$ being set to 0, resulting in a loss of information. 
It was also observed that this implementation made the resulting matrix more likely to be ill-conditioned. A heuristic approach is instead taken to address 0-valued frequencies.
In this approach, the negative inputs are made positive and scaled by the InvLIF function, and then made negative again.
This ensures the values in the objective $\hat{Y_i}$ are all similarly scaled, aiding with the solution of the resulting LSE problem.
Using this approach, the ANN activations $A_i$ may be converted into objective SNN layer inputs $\hat{Y_i}$ as shown in~\cref{eq:inlif}.  
    
    Once the objective SNN-layer inputs are known, the decoder weights may be solved for directly, which may not be feasible because many neuromorphic architectures impose constraints on neural core sizes and connectivity. 
It may be desirable, instead, to limit the size of any given SNN layer to an arbitrary size of the designer's choice. 
This is accomplished by implementing a truncation method on the target activations. 
The truncation process is outlined in~\cref{alg: trunc}. 
    
    The target matrix is first decomposed using singular value decomposition into the component $U$, $\Sigma$, and $V^T$ matrices, where $\Sigma$ is a diagonal matrix of singular values $\sigma$. 
The $N$ highest-order singular values are set to 0, where $N$ is the total number of singular values minus the target size of the SNN at that layer. 
The new singular value matrices are recombined to provide an activation matrix with the neuron dimension equal to the desired SNN layer size. 
    
    The resulting truncated matrix is a lower rank representation of the original. 
This representation is shown to be the optimal (minimum Frobenius norm difference) reduction by the Eckart-Young theorem, and may be extended to discrete integer-valued problems as well~\cite{doi:10.1137/10081099X}. 
This method is similar to existing truncation methods proposed for use on ANNs~\cite{Han2015,He2014}. 
Unlike these methods, the proposed one uses the generated sample information to truncate based on optimal hidden-layer feature representation rather than on optimal weight matrix truncation.
A disciplined study on the difference in these two methods is left for future work. 
    
    The decoders may now be solved for using the truncated activations. In this work, we solve the resulting least-squares problem using the pseudo-inverse. Using the decoders, the output of the SNN layer may be solved using the LIF rate function. 
    
    The samples used at the following hidden layer $X_{i+1}$ are then generated by adding suitable noise, $\epsilon$ to the calculated activations, as shown in~\cref{eq:end}. 
    As previously stated, the LIF rate output is only an approximation to the actual behavior of the neurons when operating in spiking mode. 
    This approximation fails to capture the temporal effects introduced by this spiking, including spike phase and synaptic filtering. 
    We treat these effects as noise in our solution by adding a noise term to the output of each neural layer carried forward in the algorithm. 
    For SNN implementation on fully-digital neuromorphic architectures, with deterministic spike interval, this noise may be neglected. 
    In this work, we sample this noise from a multi-variate Gaussian distribution, with covariance set to the covariance of the sample set scaled by the network maximum firing rate (in this work, 1000 Hz).
    
    This process is repeated for each layer in the network until the output layer. 
    At this point, since the ANN output is no longer being acted upon by an activation function, the target activations are set to be the outputs of the ANN, and the algorithm proceeds as normal. 
    The complete process is outlined in~\cref{alg: feat}. 
    
\begin{algorithm}
\caption{Layer-wise synapse optimization}\label{alg: feat}
\begin{algorithmic}[1]
\Function{Translate}{$\textit{weights}_{\text{ANN}}, \textit{Samples}, \textit{Sizes}$}
\State $W \gets \textit{weights}_{\text{ANN}}$
\State $X \gets \textit{Samples} * rep$\Comment Replicate samples
\State $S \gets \textit{Sizes}$ \Comment SNN layer sizes
\State $\Phi \gets \emptyset$
\For{$i \in n_{\textit{layers}}$}
\State $Y_i \gets \textit{Samples}*W_i$
\If{$i \not= n_{\textit{layers}}$}
\State $A_i \gets f(Y_i)$\Comment ANN activation function
\State $A_i \gets A_i*scale$
\State $\hat{Y_i} \gets \Call{InvLIF}{A_i}$
\State $\tilde{Y_i} \gets \Call{Truncate}{\hat{Y_i},S_i}$
\Else 
\State $\tilde{Y_i} \gets Y_i$
\EndIf
\State $\phi_i = X^+ \tilde{Y_i}$
\State $ X \gets \Call{LIF}{\phi_i X} + \epsilon$ \Comment Add neural noise
\State Add $\phi_i$ to $\Phi$
\EndFor
\State \Return $\Phi$
\EndFunction
\end{algorithmic}
\end{algorithm}

\begin{algorithm}
\caption{Sample Truncation}\label{alg: trunc}
\begin{algorithmic}[1]
\Function{Truncate}{$\hat{Y}, s$}
\State $U\Sigma V^T \gets \text{SVD}(\hat{Y})$
\State $\tilde{\Sigma} \gets \Sigma$
\State $\tilde{\Sigma}_{ii} \gets 0$ for $ i>s$ 
\State $\tilde{Y} \gets U\tilde{\Sigma}V^T$
\State\Return $\tilde{Y}$
\EndFunction
\end{algorithmic}
\end{algorithm}

\section{EXPERIMENTS AND RESULTS}

The feature-translation method was tested by translating three different ANNs into SNNs and evaluating the resulting networks on task performance and agreement with the ANN output. 
Two of the networks translated were multi-layer perceptron (MLP) networks. 
These both represented control policies for simulated dynamic-systems learned through reinforcement learning. 
The third network was a convolutional neural network (CNN), trained on the MNIST hand-written digit classification task~\cite{726791}. 
All the networks were trained using the Tensorflow deep learning framework~\cite{tensorflow2015-whitepaper}. 

	In order to study the effects of various translation hyper-parameters on the performance of the resulting networks, each ANN was translated under multiple parameter settings, as shown in~\cref{tab:exp_params}. 
    The integration time is the amount of time the network was allowed to accumulate charge while stimulated by the constant input signal. 
    The output signal was averaged over time to give the network output for the supplied input. ``Sample Replication" is the number of times the supplied input sample-set was replicated by the algorithm. 
    Size factor is how much each SNN layer was reduced in size, relative to the original ANN layer. 
    This size reduction was applied uniformly across all layers of the network.
	
  The resulting ANNs implement a physics-based LIF model, and therefore represent an analog or hybrid neuromorphic architecture. 
  The firing threshold voltage was set to $1\text{V}$, and each neuron was given a background current of $1\text{amp}$ to offset the resulting threshold bias.
  The neuron time constants $\tau_{m}$ were deterministically set to 0.02 with zero mismatch.
  The synaptic connections between neurons implement a simple low-pass filter, with a time constant of $1.0$ms. The networks were set to have a maximum neuron firing rate of $1000$ Hz.

\begin{table}[h]
\centering
\caption{\label{tab:exp_params}Experiment Parameters}
\begin{tabular}{ @{}p{5em}m{8em}m{6em}m{5em}@{} }
\toprule
&Integration Time & Sample \newline Replication & Size Factor\\
\midrule
Small MLP & $10$ms,\,$50$ms, $100$ms,\,$200$ms, $500$ms,\,$1000$ms & $1\times, 5\times, 10\times$ & $100\%, 75\%$, $50\%, 25\%$ \\
\midrule
Large MLP & $10$ms,\,$50$ms, $100$ms,\,$200$ms, $500$ms,\,$1000$ms & $1\times, 5\times, 10\times$ & $100\%, 75\%$, $50\%, 25\%$ \\
\midrule
CNN & $10$ms,\,$50$ms, $100$ms,\,$200$ms, $500$ms,\,$750$ms & $1\times, 5\times$ & $100\%, 75\%$, $50\%, 25\%$ \\
\bottomrule
\end{tabular}
\end{table}

All of the SNN models were built and simulated using the Nengo large-scale neural engineering system simulator~\cite{Bekolay2014}. 
Nengo is a discrete-time simulator of large scale spiking neural systems based on the Neural Engineering Framework. 
All simulations in this study were created using the built-in Nengo objects, with the exception of the neuron model.
The basic LIF neuron suffers from numerical error at high firing rates, caused by the default reset to zero voltage.
This was addressed by building a modified LIF neuron that subtracts the threshold voltage from the calculated voltage at firing time, compensating for some of the numerical discretization error. 
Without this, high-frequency signals, which are responsible for communicating a high amount of information, suffer from significant discretization loss. 

\subsection{MLP Translations} \label{sub: MLP}
\begin{figure}[!bp]
  \centering
  \begin{subfigure}[b]{0.22\textwidth}\centering
    \includegraphics[width=\textwidth]{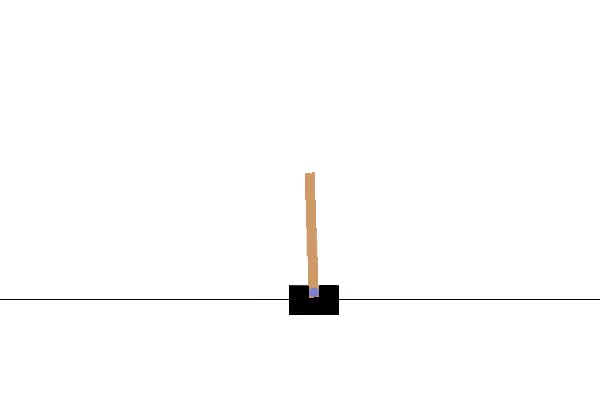}
    \caption{Cartpole\label{fig:cart}}
  \end{subfigure}
  \begin{subfigure}[b]{0.22\textwidth}\centering
    \includegraphics[width=0.7\textwidth]{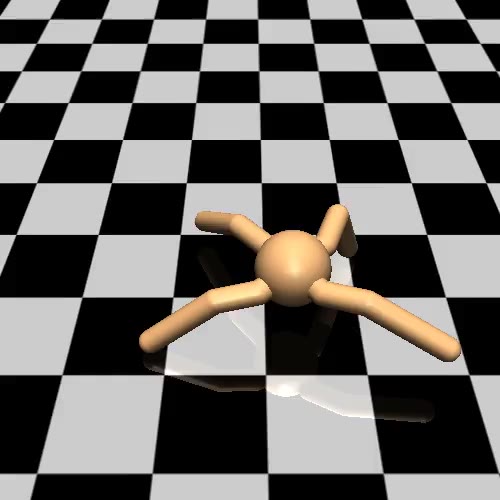}
    \caption{Ant-Walker\label{fig:ant}}
  \end{subfigure}
  \caption{Screen-shots of the OpenAI Gym environments\label{fig:envs}}
\end{figure}

Two MLPs were tested on one reinforcement learning task each from OpenAI gym~\cite{1606.01540}. Each task requires the policy to select an action from a known action space based on a provided observation.
The simulated environment is then stepped forward one discrete time-step and the process repeats with the environment providing an observation and a reward signal on each step.
The performance of a policy in on a given task is measured by the accumulation of reward signals over episodes of the task.
The first MLP (``Small MLP") was trained using stochastic gradient descent on the Cart Pole task.
The objective of this task is to keep a pole balanced as an inverted pendulum by moving the cart base in 1D.
The observation space is a four-element vector and the output is a binary action choice (``move left" or ``move right").
A screen-shot of this environment is shown in~\cref{fig:cart}.

	The translated ANN had three layers (two hidden layers), with 64 rectified linear unit (ReLU) neurons at each hidden layer. The ANN outputs unnormalized logits, which were passed to a softmax function and then used as probability masses for each discrete action. Actions were stochastically selected from the resulting distribution at each time step. During translation, the neural noise added at each layer was the clock rate, 1,000Hz scaled by the maximum standard deviation of the four input signals. Trajectories from ten episodes of 200 steps each were used as the sample inputs, for a total of 2,000 samples. The SNN was then tested over 40 episodes, for a total of $8000$ steps. The best agreement between the SNN and ANN was achieved at replication factor 5, integration time 1s, with an average RMSE across the $8000$ samples of $0.063$ with a standard deviation of $0.36$. The ANN scored an average of 200/200, with a standard deviation of 0.0. The SNN scored 200/200 with a standard deviation of 0.0. These results, along with those from the other experiments are summarized in~\cref{tab:summary}. 

	The second MLP (``Medium MLP") was trained using the Trust Region Policy Optimization (TRPO) algorithm~\cite{DBLP:journals/corr/SchulmanLMJA15} on the Ant walker task. The objective of this task is to control a four-legged robotic walker to walk as far as possible by controlling the individual joint motions. The observation space is a $111$-element vector and the output is 8 continuous valued actions (joint-torques). A screen-shot of this environment is shown in figure~\cref{fig:ant}. 

	The translated ANN had four layers (three hidden layers), with $100$, $50$, and $25$ ReLU neurons at each hidden layer. The ANN outputs the continuous, real-valued actions which are passed directly to the environment. During translation, the neural noise added at each layer was the clock rate scaled by the average standard deviation of the input signals. Trajectories from $120$ episodes of $1000$ steps each were used as the sample inputs, for a total of $120000$ samples. The best agreement with the ANN was achieved with replication factor 5, integration time 1s, with an average RMSE of $0.082$ with a standard deviation of $0.043$. The ANN scored an average of $4153.2/1000$, with a standard deviation of $796.6$. The SNN scored $4534.5/1000$ with a standard deviation of $86.1/1000$. 
\subsection{CNN Translation}

\begin{figure}
\includegraphics[width=0.5\textwidth]{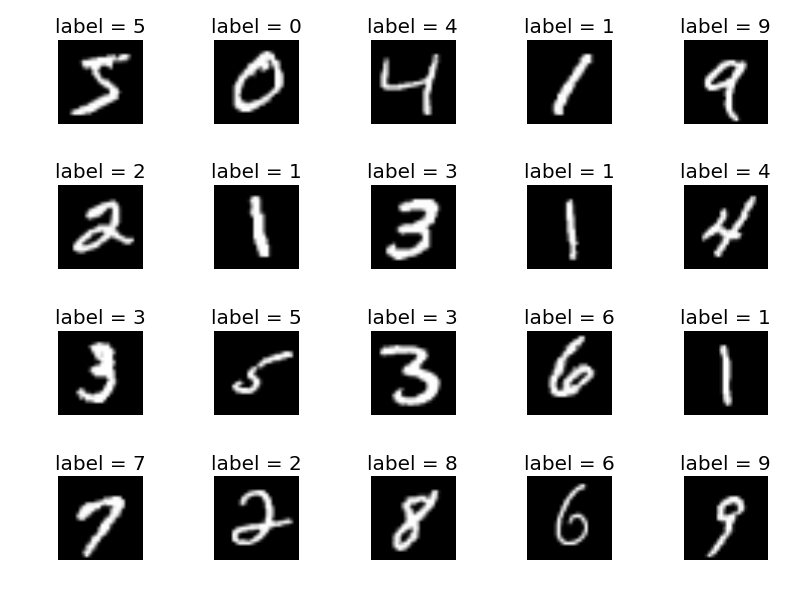}
\caption{Example of images and labels from MNIST dataset}
\label{fig:mnist}
\end{figure}

	A CNN was trained on the MNIST (\textit{mini}-NIST) dataset of hand-written digits. Each image in MNIST is a small, $28$px by $28$px, single-channel, black-and-white image of a hand-written numerical digit 0 to 9, with matched classification label. The objective of the task is to correctly identify the pictured digit. An example of the images and labels is shown in~\cref{fig:mnist}. The architecture of the CNN trained for this task is shown in the first three columns of~\cref{tab:CNN_params}. ReLU functions were used at each layer. 
    
    The CNN was trained using stochastic gradient descent with ADAM optimization~\cite{DBLP:journals/corr/KingmaB14}. The full MNIST dataset was segregated into a training and test batch, having $60,000$ and $10,000$ images respectively. No zero-centering or normalization pre-processing was done on the dataset. The single-channel input was replicated three times to create a three-channel input so that a general CNN architecture could be used. The CNN was trained on all images across 10 epochs of training using mini-batch sizes of 30 images. 
    
\begin{table}[h]
\centering
\caption{\label{tab:CNN_params}CNN Architecture}
\begin{tabular}{ @{} lcccc @{} }
\toprule
&Filter Dims. & Num. Filters & Stride&FC Dims.\\
\midrule
Conv Layer 0 & $(4\times4\times3)$ & $8$ & $4$ & $(3072\times512)$\\
Conv Layer 1 & $(4\times4\times8)$ & $16$ & $1$ & $(512\times400)$\\
Conv Layer 2 & $(4\times4\times16)$ & $32$ & $1$ & $(400\times288)$\\
Conv Layer 3 & $(4\times4\times32)$ & $64$ & $1$ & $(288\times64)$\\
FC Layer 0 & $256$ units & n/a & n/a & $(64\times256)$\\
\bottomrule
\end{tabular}
\end{table}

	This translation method, and neuromorphic architectures in general, cannot accommodate sliding convolutional layers. In neuromorphic chips, the sliding filters are often implemented in parallel cores with redundant weights for each filter location. In order to accommodate this, the CNN was transformed into a large fully-connected network by repeating the filter matrices in the appropriate configuration and reshaping the image into a column vector. The dimensions of the resulting fully-connected MLP are shown along with the original in~\cref{tab:CNN_params}.
    
    During translation, the neural noise added at each layer was the clock rate, 1,000 Hz scaled by the average standard deviation of the input pixels signals. $30,000$ images from the training set were randomly selected as the input sample-set for each translation. For testing, $1,000$ images were randomly selected from the test set. The best agreement between the SNN and CNN was achieved with no size reduction and five-time sample replication, with an average RMSE across the 1,000 samples of $7.28\%$ with a standard deviation of $4.50\%$. In this case, both the CNN and the SNN achieved an average accuracy of $100\%$ on the 1,000 samples.

\subsection{Summary Results}
\begin{figure*}
  \begin{subfigure}{0.33\textwidth}
    \includegraphics[width=\textwidth]{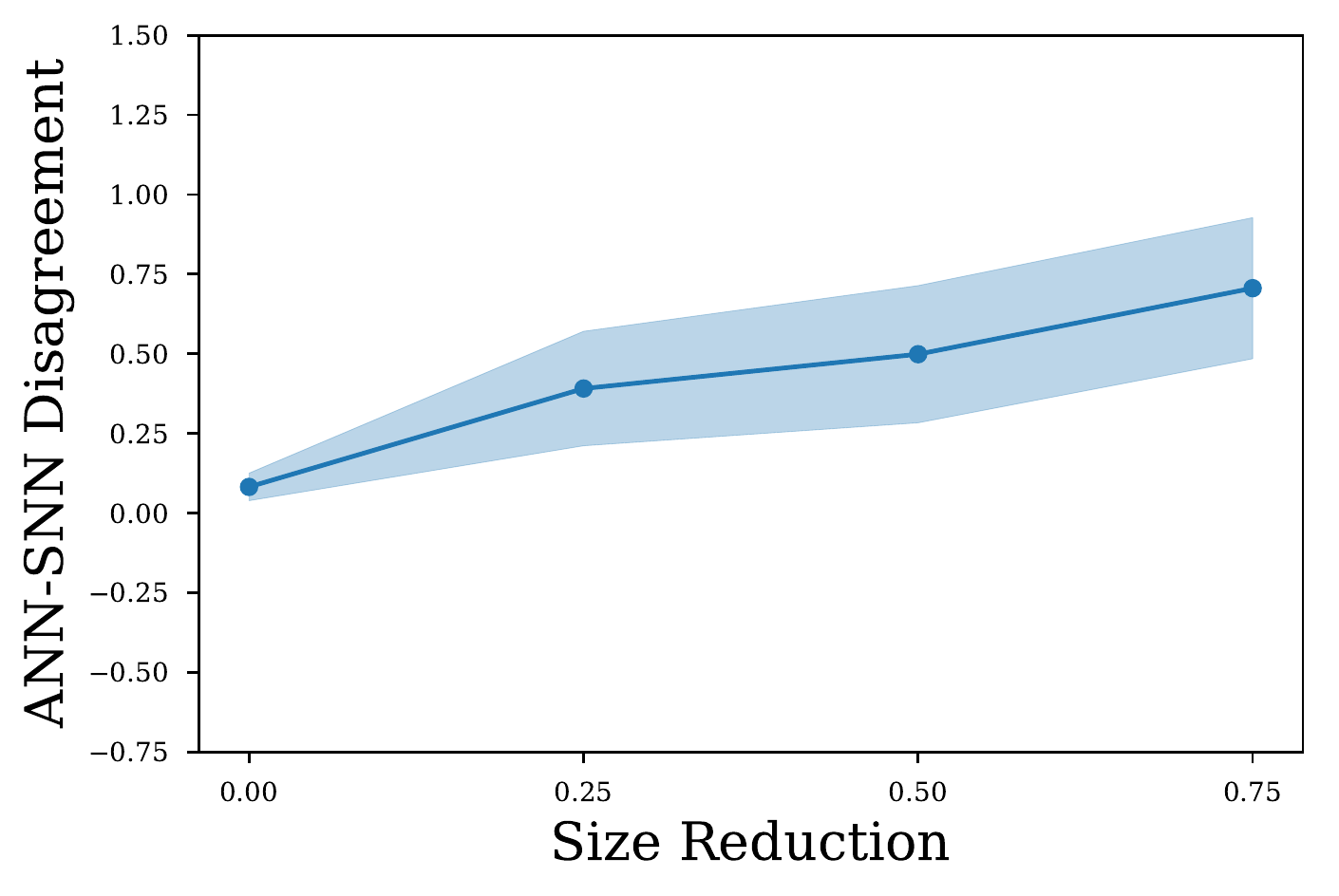}
    \caption{MLP Ant}
    \label{fig:antscale}
  \end{subfigure}
  \begin{subfigure}{0.33\textwidth}
    \includegraphics[width=\textwidth]{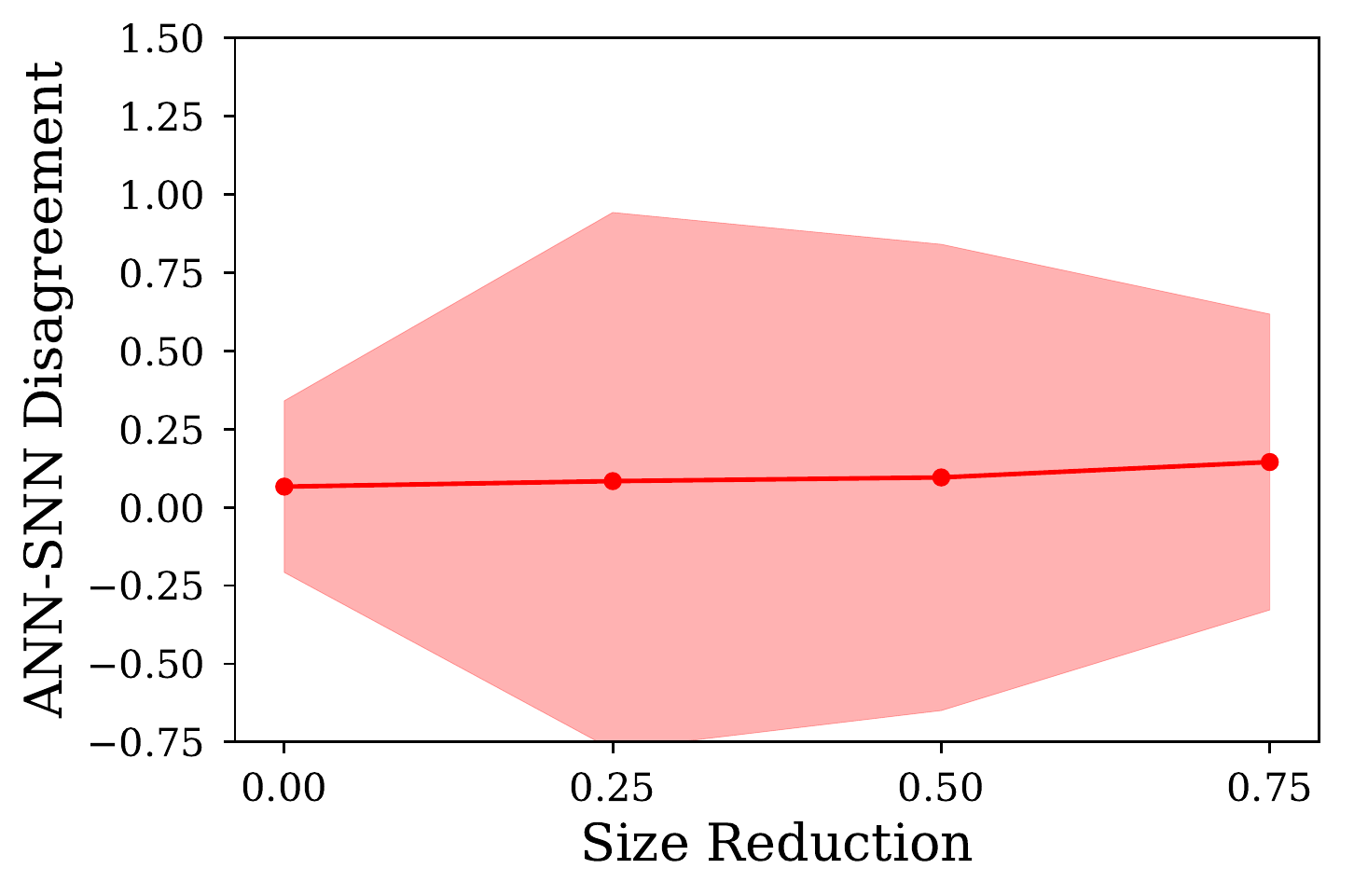}
    \caption{MLP Cartpole}
    \label{fig:cartscale}
  \end{subfigure}
  \begin{subfigure}{0.33\textwidth}
    \includegraphics[width=\textwidth]{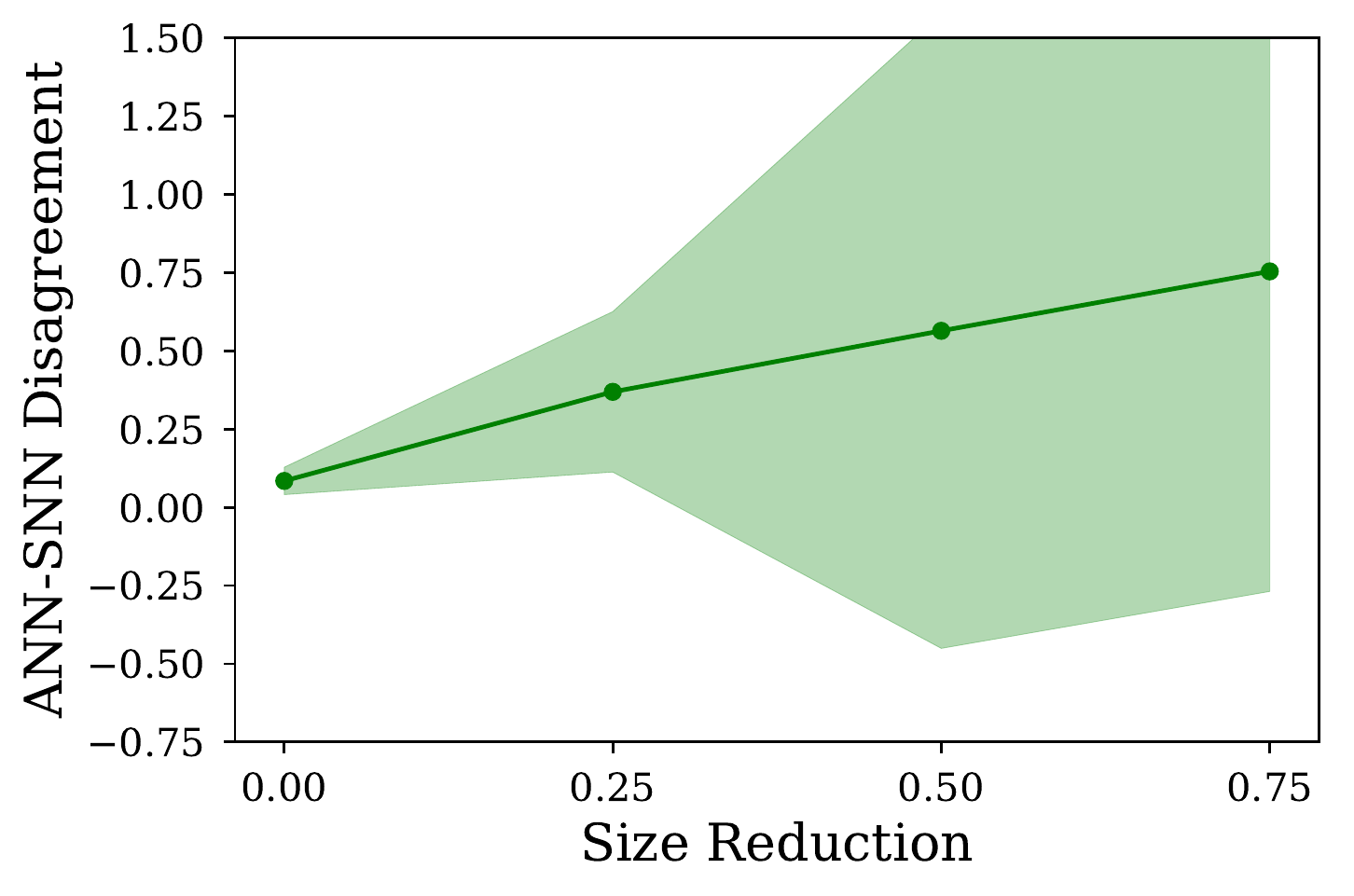}
    \caption{CNN MNIST}
    \label{fig:mnistscale}
  \end{subfigure}
  \caption{Network Disagreement vs Size Factor\label{fig:sizefact}}

\end{figure*}

The effects of the various parameters can be seen in the included figures. \Cref{fig:sizefact} shows the effect of increasing the network compression on the network disagreement. The results shown are for 500ms integration time and five-time sample replication. A strong correlation can be seen between compression amount and disagreement in both the Ant-Walker MLP and the CNN, however, there is only a slight increase in the Cart Pole MLP.

\begin{figure*}
  \begin{subfigure}{0.33\textwidth}
    \includegraphics[width=\textwidth]{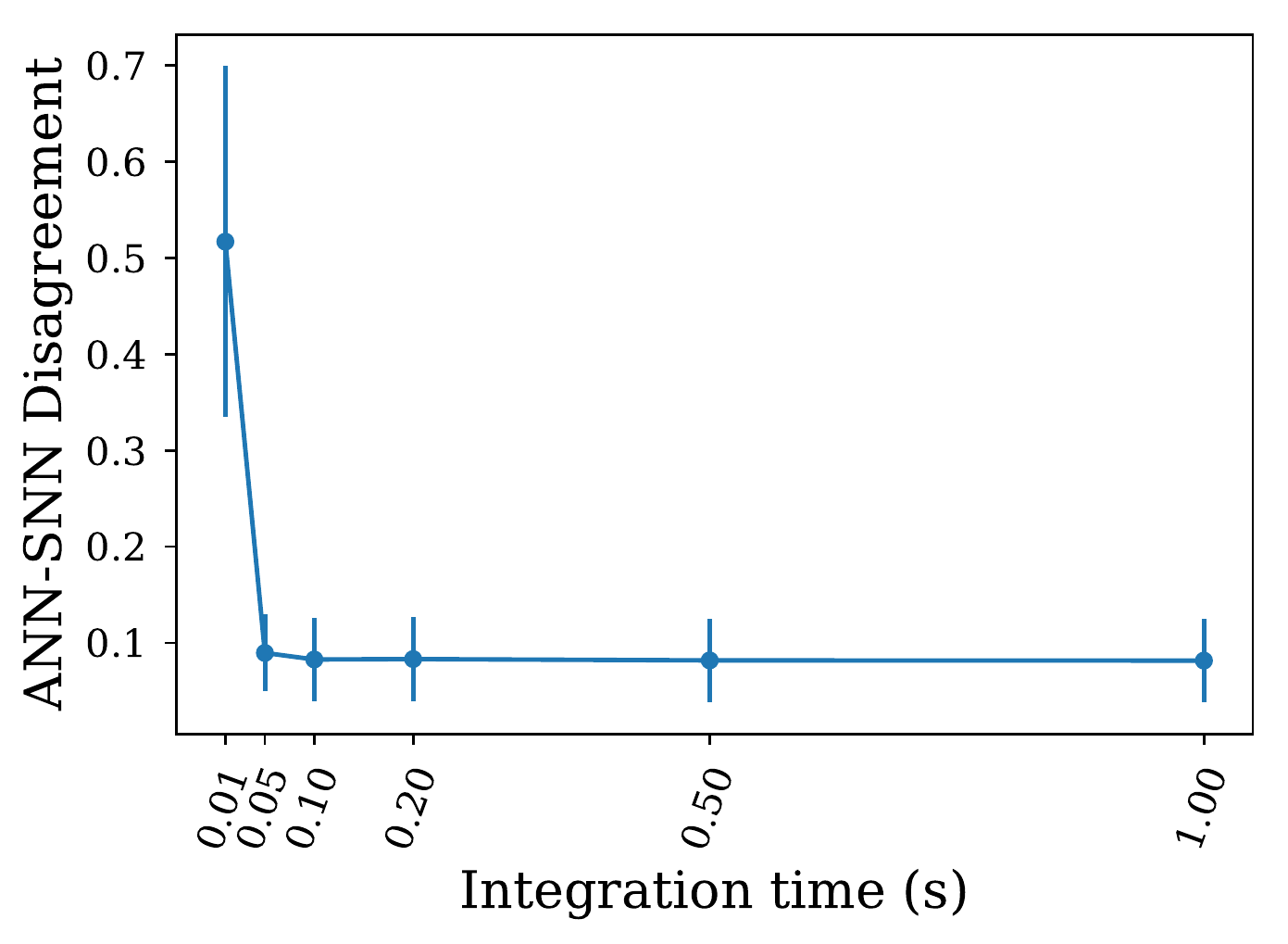}
    \caption{MLP Ant}
    \label{fig:anttime}
  \end{subfigure}
  \begin{subfigure}{0.33\textwidth}
    \includegraphics[width=\textwidth]{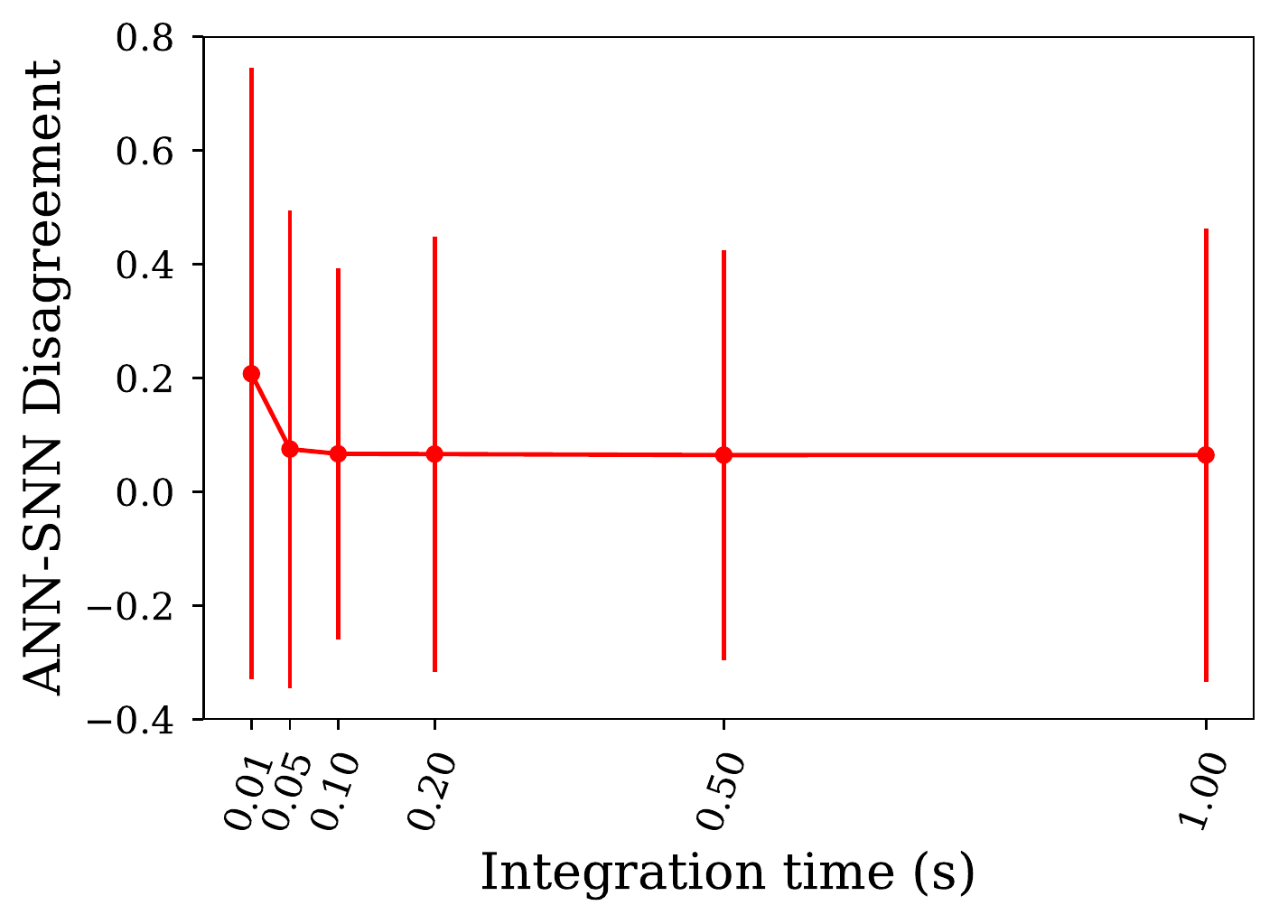}
    \caption{MLP Cartpole}
    \label{fig:carttime}
  \end{subfigure}
  \begin{subfigure}{0.33\textwidth}
    \includegraphics[width=\textwidth]{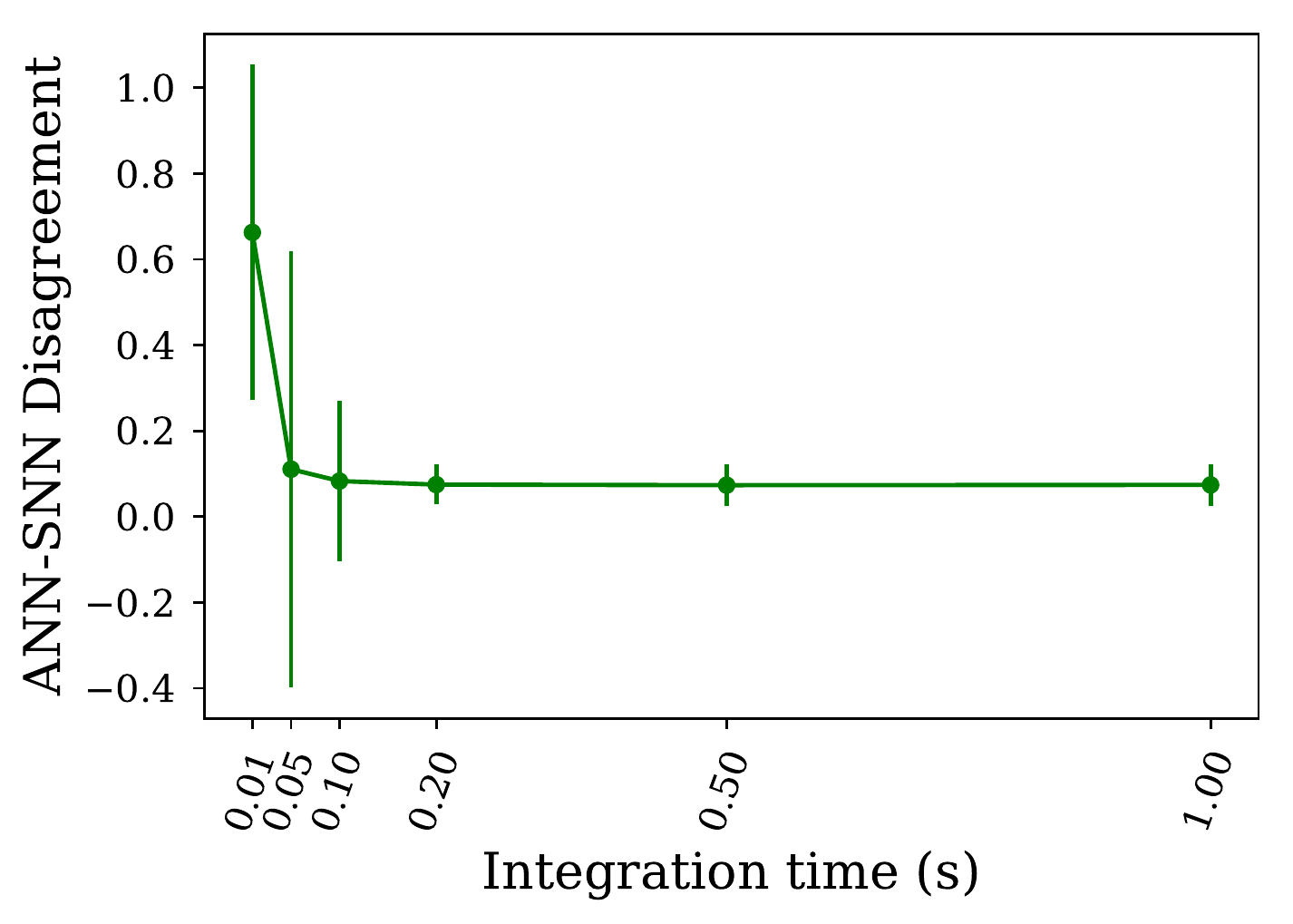}
    \caption{CNN MNIST}
    \label{fig:mnisttime}
  \end{subfigure}
  \caption{Network Disagreement vs Integration Time\label{fig:inttime}}
\end{figure*}

\Cref{fig:inttime} shows the effect of varying the SNN integration time on the relative disagreement with the ANN. The results shown are for full-sized network translation, with five-times sample replication. As can be seen, the disagreement drops sharply as integration time increase from 10ms to 50ms and then asymptotes above 100ms for all three networks. 

For the CNN, we also report the accuracy of the networks on the classification task for varying levels of compression. The results shown in \cref{tab:CNN_scale} are for 500ms integration time and five-time sample replication. The total number of neurons in the resulting SNNs are reported along with the accuracy of the SNN and the original CNN. The variance in CNN performance is due to the random sampling of the 1,000 image sub-batch used as the test set for each network instance.

\Cref{tab:summary} shows the performance of each network with differing sample replication. Each result is reported for an integration time of 500ms and a full-sized translation. For the MLP networks, the average episode score over the test runs is reported. For the CNN, the classification accuracy on the test set is reported. 

\begin{table}[h]
\centering
\caption{\label{tab:CNN_scale}CNN Performance with Scaling}
\begin{tabular}{ @{} lccc @{} }
\toprule
&No. Neurons & CNN Accuracy & SNN Accuracy \\
\midrule
100\% Size & $1520$ & $90\%$ & $92\%$\\
75\% Size & $1140$ & $98\%$ & $97\%$\\
50\% Size & $760$ & $86\%$ & $94\%$\\
25\% Size & $380$ & $100\%$ & $100\%$\\
\bottomrule
\end{tabular}
\end{table}

\begin{table}[h]
\centering
\caption{Network Performance and Disagreement}
\begin{tabular}{@{}lrS[table-format=4.1,table-figures-uncertainty=1] S[table-format=4.1,table-figures-uncertainty=1] S[table-format=1.3,table-figures-uncertainty=1]}
\toprule
Env. & {Rep.} & {ANN} & {SNN} & {Disagreement}  \\
\midrule
Ant & $1$ & 4153.2 \pm 796.6 & 4534.5 \pm 86.1 & 0.082 \pm 0.043 \\ \cmidrule{2-5}
    & $5$ & 4564.3 \pm 73.9 & 4318.7 \pm 519.4 & 0.082 \pm 0.043\\ \cmidrule{2-5}
    & $10$ & 4523.3 \pm 10.6 & 4513.0 \pm 93.2 & 0.082 \pm 0.043\\ 
\midrule
Cartpole & $1$ & 200.0 \pm 0.0 & 200.0 \pm 0.0 & 0.067\pm 0.36\\ \cmidrule{2-5}
         & $5$ & 200.0 \pm 0.0 & 200.0 \pm 0.0 & 0.063\pm 0.39\\ \cmidrule{2-5}
         & $10$ & 200.0 \pm 0.0 & 200.0 \pm 0.0 & 0.063\pm 0.39\\
\midrule
MNIST & $1$ & 0.91 & 0.91 & 0.083\pm0.019 \\ \cmidrule{2-5}
      & $5$ & 1.00 & 1.00 & 0.079\pm0.016 \\ 
\bottomrule
\label{tab:summary}
\end{tabular}
\end{table}

\section{DISCUSSION}
	The results show that layer-wise synapse optimization is able to effectively translate both MLP and CNN neural networks to SNNs compatible with implementation on neuromorphic chips with arbitrary neurons. Variance of the multiple translation parameters results in variation of the agreement between ANN and SNN.

	For all three networks, sample replication had little effect on either network agreement or task performance. This may be because for all three networks, relatively large sample sets were used which provided redundant coverage of the input space and enough capacity to obviate the need for sample replication. Sample replication may remain a viable method to increase translation robustness for tasks in which sample data is sparse or generation of samples is expensive. Investigation of this hypothesis is left for future work. 
    
    In several domains, particularly those involving system control, integration time may be a critical performance parameter. The networks produced in this study show were able to retain agreement and task performance for integration times as low as 50ms for a 1,000 Hz system. In general, there is an input-output delay in time-dependent SNNs. In this work, outputs from the first 10\% of the total integration time were ignored to account for this delay. This may have resulted in too small a delay window for the low integration-time runs. For task-specific networks, a measurement of the delay could be completed to more precisely set the window and allow for smaller integration times. 

	Network compression had a significant effect on network agreement for the ant-walker MLP and the CNN. For both of these networks, average disagreement increased by approximately 60\% as network size was decreased from 100\% of the original to just 25\%. For the Cart Pole MLP, little effect was observed with decrease in layer size, perhaps because the original ANN over-parameterized the relatively simple problem, resulting in very sparse weights. It should be noted, that although the CNN agreement degraded significantly, the task performance did not, with the SNN meeting within one percent or exceeding the CNN accuracy in for each test case. This suggests that the disagreement observed was the effect of approximately uniform scaling of the outputs, retaining the critical relative relationship between output magnitudes required for classification. The fact that the translated CNN performance improves with increasing compression may be a result of the flattening process used to convert the CNN into a densely connected MLP. This conversion results in large, sparse weight matrices that can be ill-conditioned. The proposed truncation method removes the higher-order singular values, resulting in a lower condition number.

	In this work, the same network compression factor was used on every layer in each test case. This may be improved upon by instead applying differing compression factors to each layer. In general, compression works better on networks with sparse weights or with most weights near zero~\cite{Han2015,He2014}. Therefore, by applying higher compression to these types of layers, and less to more dense layers, more effective compression can be achieved with lower loss in accuracy. 
    
    This compression would be especially beneficial for implementation of CNNs. As shown, the expansion of the sliding convolutional layers to fully-connected increased the overall parameter count per-layer, for example from 384 to over 1.5 million for the first layer of our network. Deploying this to any practical neuromorphic chip would be infeasible. At the same time, the method used to convert the convolutional layers to fully-connected results in very sparse weights. For example, for the CNN in this work $98\%$, $75\%$, and $64\%$ of the weights from the fully-connected convolutional layer conversions are zero. This result suggests that network compression, when applied  to these layers, would be an effective method to compress CNNs for practical neuromorphic deployment.

\section{Conclusions}

	In this work we presented layer-wise synapse optimization (LSO), a method to convert ANNs layer-by-layer by optimal representation of the network hidden-layer activations. The method may be used to convert ANNs to an SNN with physically realistic neurons. This offers a significant improvement on previous methods, for which limitations on both ANN activations and SNN neurons limited their utility to only digital neuromorphic architectures. In contrast, feature-translation can generate SNNs compatible with digital, analog and hybrid architectures. LSO was shown to be effective on MLP networks and, with appropriate conversion to a dense MLP, convolutional networks. For all the networks tested, good agreement was achieved between the SNN and the original ANN. The networks translated represented policies for two distinct classes of problems: reinforcement learning-based system control and image classification. LSO was able to maintain excellent performance across both of these task types. Initial results also show that small integration times, on the order of 50ms for a 1,000 Hz system, are attainable. 
    
	LSO also introduces an optimal compression method for ANN conversion. In contrast to previous methods which often require replication of ANN nodes, this method allows the network size to be selected by the designer by optimally compressing the ANN layers. This is critical for deployment of the translated SNNs to practical neuromorphic systems which often impose constraints on core size and connectivity. As seen, conversion of CNNs to MLPs results iton extremely large networks that would be infeasible to deploy on many systems. Through use of the novel compression method introduced, these layers can be truncated to manageable dimensions. In this work, compression to a network as small as $25\%$ of the original was accomplished with no loss in performance on the MNIST task. 
    
    There are several extensions to this method that are left for future work. Additional spiking neuron types should be investigated. In particular, SNNs representative of digital neuromorphic systems should be generated. End-to-end validation could then be conducted by deploying and testing the network on a neuromorphic chip. Our primary choice for this test is the TrueNorth chip from IBM~\cite{Merolla2014a}. Deployment to chip would also require incorporation of the particular constraints imposed by the hardware. For TrueNorth, these are primarily constraints on the resolution of the synaptic weights and on neuron connectedness. These can be addressed through use of a constrained optimizer to solve the LSE problem in the layer translation. 
    
    During the study, it was observed that the large matrices generated resulted in high memory cost and slow performance. An alternative solver to the matrix inversion method should be investigated to alleviate this computational cost. Further investigation should also be done to study the effect of network depth on the effectiveness of the translation method. Lossy errors have a tendency to accumulate through SNN layers~\cite{Rueckauer2016}; hence, the robustness of our method should be studied.
    
    LSO is based on approximating neuron spiking behavior as continuous firing-rate curves. Additional performance may be gained by optimizing network performance directly on the spiking-network after training, to fine-tune the weights. Investigation of evolutionary or cross-entropy based approaches for this fine-tuning should be conducted. Additionally, differentiable ANN activation functions that approximate neuromorphic firing rate curves should be investigated~\cite{Hunsberger2016}. These activation functions would allow for gradient-based training such that the resulting ANN weights would be closer to the weights required for SNN implementation and would likely improve the accuracy of the final network after translation. 

\addtolength{\textheight}{-10cm}   



\section*{ACKNOWLEDGMENT}
The authors would like to thank the U.S. Army Research Laboratory's High Performance Computing Research Center for the support of this work, especially Manuel Vindiola and Vinnie Monaco. The authors are also grateful to Professor Kwabena Boahen and Eric Kauderer-Abrams from the Stanford Neuromorphics Laboratory for their guidance and advice. 

\bibliographystyle{ieeetr}
\bibliography{bib2}


\end{document}